\documentclass[letterpaper, 10 pt, conference]{ieeeconf}
\IEEEoverridecommandlockouts

\usepackage{cite}
\usepackage{amsmath,amssymb,amsfonts}
\usepackage{algorithmic}
\usepackage{graphicx}
\usepackage{textcomp}
\usepackage{mathtools}
\usepackage{amsmath}
\usepackage{mhchem}
\usepackage{tensor}
\usepackage{gensymb}
 \PassOptionsToPackage{hyphens}{url}\usepackage{hyperref} % https://tex.stackexchange.com/questions/3033/forcing-linebreaks-in-url
\usepackage[dvipsnames]{xcolor}
\usepackage{listings,multicol}
\title{\LARGE \bf
Whole-Body Control on Non-holonomic Mobile Manipulation for Grapevine Winter Pruning Automation*
}

\author{Tao Teng$^{1,3}$, Miguel Fernandes$^{2}$,  Matteo Gatti$^{3}$, Stefano Poni$^{3}$,  Claudio Semini$^{1},$ \textit{Senior Member, IEEE}, \\ Darwin Caldwell$^{2}$, \textit{Senior Member, IEEE}, Fei Chen$^{\dagger  4 }$, \textit{Senior Member, IEEE}% <-this % stops a space
\thanks{*This study was supported by the Doctoral School on the Agro-Food System (Agrisystem) of Universit\`a Cattolica del Sacro Cuore (Italy), and was supported in part by the project “Grape Vine Perception and Winter Pruning Automation” funded by joint lab of Istituto Italiano di Tecnologia and Universit\`a Cattolica del Sacro Cuore, and the project “Improving Reproducibility in Learning Physical Manipulation Skills with Simulators Using Realistic Variations” funded by EU H2020 ERA-Net Chist-Era program.}% <-this % stops a space
\thanks{$^{1}$Dynamic Legged Systems  (DLS) Lab, Istituto Italiano di Tecnologia,
Via S. Quirico 19D, 16163 Genoa, Italy (e-mail: name.surname@iit.it)}%
\thanks{$^{2}$Department of Advanced Robotics, Istituto Italiano di Tecnologia, Via S. Quirico 19D, 16163 Genoa, Italy (e-mail:name.surname@iit.it)}%
\thanks{$^{3}$Department of Sustainable Crop Production, Universit\`a  Cattolica del Sacro Cuore, Via Emilia Parmense 84, 29122 Piacenza, Italy (e-mail: name.surname@unicatt.it)}
\thanks{$^{4}$Department of Mechanical and Automation Engineering, T-Stone Robotics Institute, The Chinese University of Hong Kong, Hong Kong (e-mail: f.chen@ieee.org)}%
\thanks{$\dagger$Corresponding author}%
}
\begin{document}

\maketitle  

%\title{Whole-Body Control on Non-holonomic Mobile Manipulation for Grapevine Winter Pruning Automation}

\begin{abstract}
Mobile manipulators that combine mobility and manipulability, are increasingly being used for various unstructured application scenarios in the field, e.g. vineyards. Therefore, coordinated motion of the mobile base and manipulator is an essential feature of the overall performance. In this paper, we explore a whole-body motion controller of a robot which is composed of a 2-DoFs non-holonomic wheeled mobile base with a 7-DoFs manipulator (non-holonomic wheeled mobile manipulator, NWMM) This robotic platform is designed to efficiently undertake complex grapevine pruning tasks. In the control framework, a task priority coordinated motion of the NWMM is guaranteed. Lower-priority tasks are projected into the null space of the top-priority tasks so that higher-priority tasks are completed without interruption from lower-priority tasks. The proposed controller was evaluated in a grapevine spur pruning experiment scenario.
\end{abstract}

%\begin{IEEEkeywords}
%	Whole-body Control, Non-holonomic Mobile Manipulator, Stack of Tasks
%\end{IEEEkeywords}
%\renewcommand{\thefootnote}{}
%\footnotetext{}
%\footnote{1. Department of Dynamic Legged Systems, Istituto Italiano di Tecnologia,
%Via S. Quirico 19D, 16163 Genoa, Italy (e-mail: name.surname@iit.it)}
%\footnote{2. Department of Advanced Robotics, Istituto Italiano di Tecnologia, Via S. Quirico 19D, 16163 Genoa, Italy (e-mail:name.surname@iit.it)}
%\footnote{3. Department of Sustainable Crop Production, Universit\`a  Cattolica del Sacro Cuore, Via Emilia Parmense 84, 29122 Piacenza, Italy (e-mail: name.surname@unicatt.it)}
%\footnote{4. Department of Mechanical and Automation Engineering, T-Stone Robotics Institute, The Chinese University of Hong Kong, Hong Kong (e-mail: f.chen@ieee.org)}
%
%\footnote{*Correspond author.}

\section{Introduction}

Technology is one of the key driving forces in the development of precision agriculture \cite{mcbratney2005future,stafford2000implementing,zhang2002precision}. Robots are incredibly effective in combating the pressures of population growth \cite{vougioukas2019agricultural}. Agricultural robots are not only able to help farmers solve labor shortages, but they also help to mitigate environmental impact \cite{kumar2020review}. In the drive to increase yields and maximize resources through the utilization of new technologies, grapevine pruning automation is a typical application scenario in precision agriculture \cite{fourie2021towards,botterill2017robot}.  The main purpose of grapevine winter pruning is to determine the number and location of the nodes remaining over the winter. These remaining nodes will grow into new canes in the next harvest season, and grapes will grow on these new canes. As a consequence, it fundamentally determines the final yield. Such tasks can be accomplished with a mobile manipulator extends the manipulability and operational space by combining the strengths of the mobile base and manipulator.

\begin{figure}[!htb] %H为当前位置，!htb为忽略美学标准，htbp为浮动图形
\centering %图片居中
\includegraphics[width=0.5\textwidth]{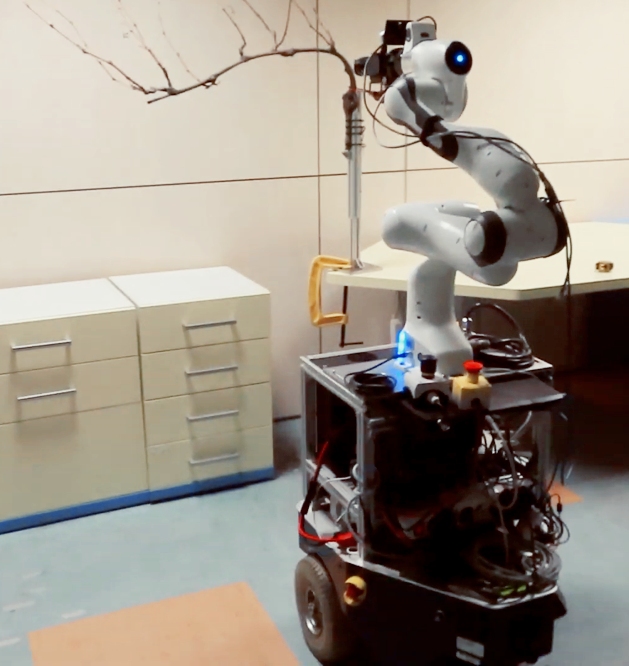} %插入图片，[]中设置图片大小，{}中是图片文件名
\caption{ An example of a robot grapevine pruning system Rolling Panda. To achieve this scenario, the following temporal sequences are followed: grapevine spur detection, whole-body motion to approach pruning point, and cutting the spur.} %最终文档中希望显示的图片标题
\label{Fig.main2} %用于文内引用的标签
\end{figure}
This combined, more complex structure creates motion planning and control issues due to the high redundancy and versatility of mobile platform \cite{raja2019learning}.  Previously, the mobile base and manipulator could be considered to be two separate subsystems \cite{tsay2010material,chen2018dexterous,10.1007/978-3-030-71356-0_1,wrock2011decoupled}. Wrock et al. created an automatic switching scheme that used teleoperation methods to achieve decoupled mobile manipulator motion. Decoupled motion refers to motion in which the mobile base remains fixed while the manipulator is operated, or vice versa. This provides high tracking accuracy at the end effector, but it takes more time to complete the task because tracking must be interrupted during motion. Coordinated movement of the mobile base and manipulator \cite{kim2019whole,hamner2010autonomous} is required to fully utilize the  strengths of mobile base and manipulator and boost efficiency.

De Luca et al. \cite{de2006kinematic} proposed a comprehensive theory to deal with modeling and redundancy resolution for non-holonomic mobile manipulators. A  multi-tasks whole-body regulating strategy based on velocity control was proposed by Li et al \cite{li2020design} 
for a highly redundant mobile manipulator. The mobile manipulator follows the predefined end-effector  trajectory while avoiding low-priority control primitives by using null-space projection. To obtain the inverse kinematics, Roberto et al. \cite{ancona2017redundancy} provided a systematic mobile manipulator solution that included a selection of redundancy parameters. The solution was capable of managing obstacle avoidance, mobile base motion restriction, and dexterity enhancement.

In this paper, a whole-body motion controller enhanced by hierarchical tasks that can regulate a non-holonomic mobile manipulator for grapevine winter pruning is proposed. The trajectory of the end effector, which is treated as the top-priority task, is created by using quintic polynomial programming. Conflicts between the end-effector tasks and the constraint tasks are handled within the stack-of-tasks (SoTs) framework\cite{katyara2021SoT,rocchi2015opensot,katyara2021intuitive}  by correctly assigning an order of priorities to the given tasks and then ensuring  that the lower priority tasks are projected into the null space of the higher-priority tasks.

The manuscript is organized in the following way: Firstly, Sec. II introduces the overall non-holonomic wheeled mobile manipulator system platform {Rolling Panda}. Sec. III presents the kinematics and dynamics models of this NWMM system. Whole-body controller for the non-holonomic mobile manipulator is designed in Sec. IV.  Subsequently, Sec. V. describes the experimental validations of the proposed control scheme. Finally, the paper is concluded in Sec. VI.

\section{Overview of Non-holonomic  Mobile Manipulator System}

The designed grapevine winter pruning system is called {Rolling Panda}. It primarily consists of a non-holonomic wheeled mobile base, a manipulator, an RGB-D (Intel Realsense D435i) eye-in-hand camera, and  pruning clippers.
\subsection{Hardware of the Mobile Manipulator System}

%\begin{equation}
% 	\prescript{14}{2}{\mathbf{C}}	
%\end{equation}

As shown in Fig. 1, {Rolling Panda} consists of two parts: the velocity controlled two-wheel non-holonomic mobile robot, MP-500 (Neobotix GmbH. Co.) and 7-DoFs robot arm manipulator, Panda (Franka Emika. Co.). Both the mobile base and manipulator have their own controller interfaces which are simple and user-friendly programming interfaces. On this mobile base, there is a  ROS interface for  low-level, real-time velocity controller and localization algorithms using wheel \footnote{ https://robots.ros.org/neobotix-mp-500/ }. The localization algorithm returns the direction and position and velocity of the mobile base's central frame in relation to its global frame. The Franka ROS Interface provides utilities for controlling and managing the Franka Emika Panda \footnote{  https://frankaemika.github.io/docs/ }. The control frequencies of the manipulator and the base are 1 kHz and 50 Hz, respectively. The ROS master laptop, used for the controller, is a core-i7 processor 1.8 GHz with 16 GB RAM.  The RGB-D camera (Intel Realse D435i) is mounted on the end effector of the {Rolling Panda}.

%On the other hand, Fig. 7 shows our system structure with the mobile manipulator. Because our mobile base supports only the velocity controller, we applied the well-known admittance control law [7] to transfer the desired torque obtained the proposed controller to the desired velocities for the mobile base.

\section{Non-holonomic Mobile Manipulator Modeling}

In this section, we explored a system kinematic and dynamic modeling for the non-holonomic mobile manipulator Polling Panda.

\subsection{Kinematic Modeling and Kinematic Constraints}
Considering the non-holonomic mobile manipulator with $2$ wheels and $7$-DoFs manipulator. The moving base of the {Rolling Panda} contains 3-DoFs of rigid body motion, hence a generalized coordinate system can be defined as following:

\begin{figure}[t] %H为当前位置，!htb为忽略美学标准，htbp为浮动图形
\centering %图片居中
\includegraphics[width=0.5\textwidth]{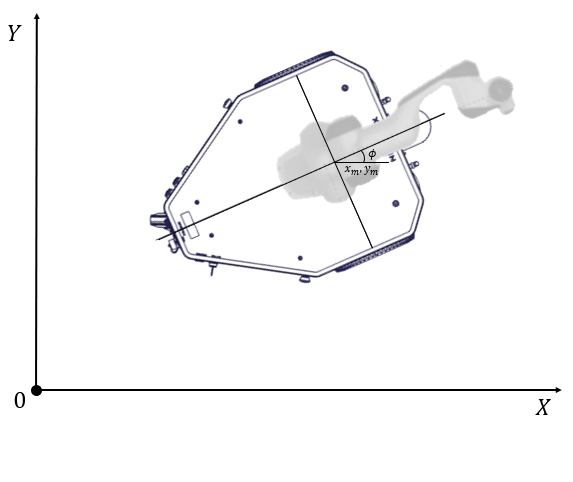} %插入图片，[]中设置图片大小，{}中是图片文件名
\caption{Schematic drawing of {Rolling Panda}. $x_m$, $y_m$, and $\phi$ denote the center position and heading angle of the mobile platform with respect to the global coordinate.} %最终文档中希望显示的图片标题
\label{Fig.main2} %用于文内引用的标签
\end{figure}

\begin{equation}
q=\left[\begin{array}{lll}
q_{m}^{T} & q_{w}^{T} & q_{n}^{T}
\end{array}\right]^{T}
\end{equation}
where $q_{m}=\left[\begin{array}{lll}x_{m} & y_{m} & \phi\end{array}\right]^{T} \in \mathbb{R}^{3}$ is the coordinate of the rotation central frame of the mobile base, $q_{w}=\left[\begin{array}{ll}\theta_{l} & \theta_{r}\end{array}\right]^{T} \in \mathbb{R}^{2}$ is spinning of the wheel joints and $q_{n} \in \mathbb{R}^{n}$ is the joint vector for the manipulator. $\mu$ is the distance between the driving wheels and the mobile platform geometric center , $\rho$ is the distance from the mobile platform rotation center to the center of mass of mobile platform, and $R$ is the radius of the wheels.

Due to its inherent properties a differential mobile platform can not move sideways \cite{liu2021whole}.
Hence,  the velocity of the mobile base in the lateral directions should be zero.

\begin{equation}
-\dot{x}_{m} \sin \phi+\dot{y}_{m} \cos \phi - \rho \dot{\phi}=0
\end{equation}

The other two constraints are a pure rolling constraint, relating the base velocities $\dot{x}_{m}, \dot{y}_{m}, \dot{\phi}$ while the wheel velocities $\dot{\theta}_{l}, \dot{\theta}_{r},$ ensure the no-slip condition at each rolling wheel in the forward directions.

\begin{equation}\begin{aligned}
\dot{x}_{m} \cos \phi+\dot{y}_{m} \sin \phi - \mu \dot{\phi} &=R \dot{\theta}_{l} \\
\dot{x}_{m} \cos \phi+\dot{y}_{m} \sin \phi + \mu \dot{\phi} &=R \dot{\theta}_{r} 
\end{aligned}\end{equation}

We can set the constraint matrix between rigid body motion of the mobile base and the generalized coordinates to satisfy the following equation.
\begin{equation}
A(q) \dot{q}=0
\end{equation}
where $A(q) \in \mathbb{R}^{3 \times\left(5+n\right)}$ is the full-ranked constraint matrix. 

\begin{equation}A=\left[\begin{array}{ccccccc}
-\sin \phi & \cos \phi & -\rho & 0 & 0 & \cdots & 0 \\
-\cos \phi & -\sin \phi & -\mu & R & 0 & \cdots & 0 \\
-\cos \phi & -\sin \phi & \mu & 0 & R & \cdots & 0
\end{array}\right]\end{equation}

 Using the null-space of $A(q),$ we can obtain the following transform equation,

\begin{equation}\label{zeroJaco}
\dot{q}=S(q) \dot{\xi}
\end{equation}
where $S(q)$ satisfies $A(q)S(q) = 0$, and 
{\small
\begin{equation}
\begin{split}
\begin{array}{c}
S= {\left[\begin{array}{ccccc}
c(\mu \cos \phi-\rho \sin \phi) & c(\mu \cos \phi+\rho \sin \phi) & 0 & \cdots & 0 \\
c(\mu \sin \phi+\rho \cos \phi) & c(\mu \sin \phi-\rho \cos \phi) & 0 & \cdots & 0 \\
c & -c & 0 & \cdots & 0 \\
1 & 0 & 0 & \cdots & 0 \\
0 & 1 & 0 & \cdots & 0 \\
0 & 0 & 1 & \cdots & 0 \\
\vdots & \vdots & \vdots & \ddots & \vdots \\
0 & 0 & 0 & \cdots & 1
\end{array}\right]}
\end{array}
\end{split}
\end{equation}
}where $c = R/\left( 2\mu \right) $. The set of feasible velocities may be expressed in terms of a suitable vector, $\dot{\xi}=\left[\begin{array}{cc}\dot{q}_{w}^{T} & \dot{q}_{n}^{T}\end{array}\right]^{T} \in \mathbb{R}^{2+n}$ is the joint velocity for actuators of the robot. 
%$\dot{\underline{z}}_{n \times 1}=\left[\dot{\theta}_{R} \dot{\theta}_{L} \dot{\theta}_{1} \dot{\theta}_{2}\right]^{\mathrm{T}}$ as

Jacobian matrix \cite{zhao2019design} between Cartesian velocity space and actual joint velocity space, $J_{\xi} \in \mathbb{R}^{6 \times(2+n)},$ can be derived as

\begin{equation}
J_{\xi}(q)=J_q(q) S(q)
\end{equation}
where $J_q(q)$ is the Jacobian matrix for $q$.

\begin{figure}[!htb] %H为当前位置，!htb为忽略美学标准，htbp为浮动图形
\centering %图片居中
\includegraphics[width=0.5\textwidth]{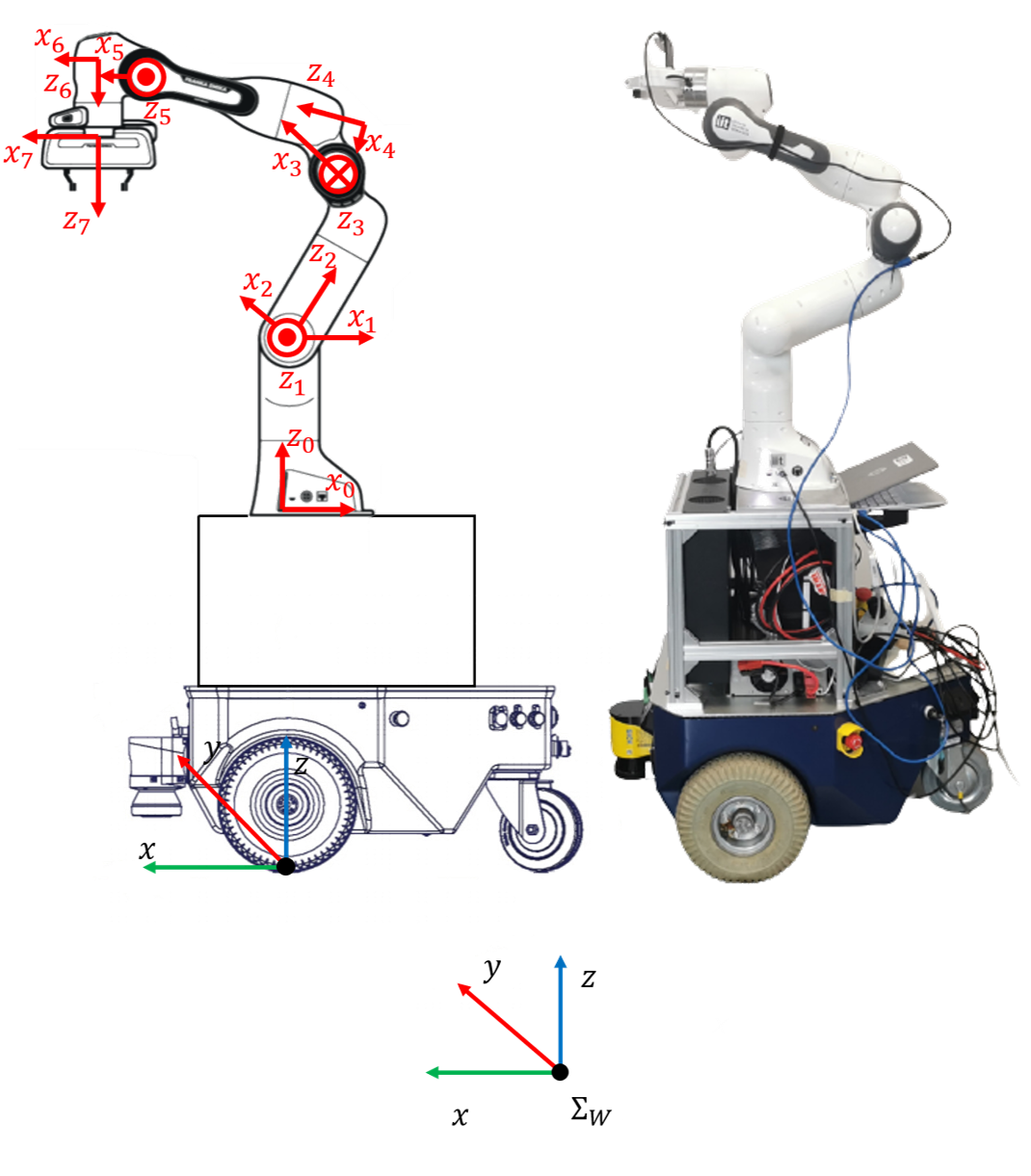} %插入图片，[]中设置图片大小，{}中是图片文件名
\caption{Whole-body kinematic model of Rolling Panda} %最终文档中希望显示的图片标题
\label{Fig.main2} %用于文内引用的标签
\end{figure}

%\begin{equation}
%\begin{split}
%S=\left[
%\begin{array}{ccccc}
%c(b \cos \phi-d \sin \phi) & c(b \cos \phi+d \sin \phi) & 0 & \cdots &0\\
%c(b \sin \phi+d \cos \phi) & c(b \sin \phi-d \cos \phi) & 0 & \cdots &0\\
%c & -c & 0 & \cdots &0\\
%1 & 0 & 0 & \cdots  &0\\
%0 & 0& 1 & \cdots  &0\\
%\vdots & \vdots & \vdots & \ddots &0 \\
%0 & 0 & 0 & \cdots &1\\
%\end{array}\right]
%\end{split}
%\end{equation}

\subsection{Dynamic model}
The following is the unconstrained equation of motion for a non-holonomic handheld manipulator \cite{kim2019whole}:
\begin{equation}\label{undynamic}
M(q) \ddot{q}+V(q, \dot{q}) \dot{q} + G(q)=B(q) \tau + \tau_{dis} +\Lambda^{T}(q){\lambda}
\end{equation}
where $M(q)$ is an $n\times n$ symmetric positive definite inertia matrix, $V(q, \dot{q})$ is the centripetal and coriolis matrix, $G(q)$ is the gravitational vector, $\tau_{dis}$ is the vector of bounded unknown disturbances including unstructured unmodeled dynamics, $B(q)$ is the input matrix, $\tau$ is the input torque vector, $\Lambda^{T}(q)$ is the matrix associated with the kinematic constraints, and $\lambda$ is the Lagrange multipliers vector.

Finally, the dynamic motion of no-holonomic mobile manipulator with respect to $\xi$ and $\dot{\xi}$ can be reformulated by removing the generalized constrains, $\Lambda(q)^{T} \lambda,$ in \eqref{undynamic} by using \eqref{zeroJaco} and combining \eqref{undynamic} and time derivative of \eqref{zeroJaco}, as follow:

\begin{equation}
{M}_{\xi}(q) \dot{\xi}+ {V}_{\xi}(q, \dot{q}) \xi + {G}_{\xi}(q)=u+S(q)^{T} \tau_{dis}
\end{equation}
where
$$
\begin{array}{l}
{M}_{\xi}(q)=S(q)^{T} M(q) S(q) \\
 {V}_{\xi}(q, \dot{q})=S(q)^{T}[M(q) \dot{S}(q)+C(q, \dot{q}) S(q)] \\
{G}_{\xi}=S(q)^{T} G(q) \\
u=S(q)^{T} B(q) \tau.
\end{array}
$$

%Hereafter, to simplify notations, $\tilde{M}$, $\tilde{V}$, $\tilde{G}$, and ${J}_{\xi}$ denote $\tilde{M}(q)$, $\tilde{V}(q, \dot{q})$, $\tilde{G}(q)$, and ${J}_{\xi}(q)$, respectively.

\section{PROPOSED CONTROLLER DESIGN}

To automate the  grapevine winter pruning, a hierarchical control formulation based on  multi-tasks scheduling has been designed.

%
%\begin{figure}[!htb] %H为当前位置，!htb为忽略美学标准，htbp为浮动图形
%\centering %图片居中
%\includegraphics[width=0.5\textwidth]{figures/Systemstructure.png} %插入图片，[]中设置图片大小，{}中是图片文件名
%\caption{whole-body control diagram} %最终文档中希望显示的图片标题
%\label{Fig.main2} %用于文内引用的标签
%\end{figure}

\subsection{ Null-Space Dynamic Control Strategy}

Robots with redundancy  (particularly high redundancy)  can deal very effectively with  constrained tasks in the Cartesian space. The redundant self motion can can simultaneously satisfy both, the higher-priority end-effector task and the additional  low-priority constrained tasks.

The task-space augmentation principle incorporates a constraint task that must be performed simultaneously with the end effector task. In this case, an augmented Jacobian matrix is constructed, the inverse of which yields the required joint velocity solution \cite{slotine1991general}. 

The relation between the $i$-th configuration  coordinate vector $\xi_i$ and the $i$-th Cartesian space task vector ${x}_{\mathrm{i}}$ can be considered as a direct kinematics equation:

\begin{equation}
\dot{{x}}_{\mathrm{c}}={J}_{\xi} \dot{\xi}
\end{equation}

The well-known generalized Moore-Penrose pseudo inverse $J^+(q)$ is used since the inverse of the nonsquare (analytical) Jacobian  $J_{\xi}$ does not exist in the redundant case. This proposed strategy often employs a special solution of equation (2).

Optimization criteria for the redundant self motion can be supplemented by, for example, null-space projection, which leads to the relation:
\begin{equation}
\dot{\xi}={J}_{\xi}^{+} \dot{{x}}_{\xi}+\left({I}-{J}_{\xi}^{+} {J}_{\xi}\right) \dot{\xi}_{0}
\end{equation}

The expression $({I}-{J}_{\xi}^{+} {J}_{\xi})$ represents the orthogonal projection matrix in the null space of $J_{\xi}$, and $\dot{\xi}_{0}$ is an arbitrary joint-space velocity satisfying argumented constraint tasks; hence, the second part of the solution is therefore a null-space velocity.

\section{GRAPEVINE WINTER PRUNING EXPERIMENTS}

Grapevine winter pruning is a key agricultural activity involving  cutting of the canes of each vine that grew during the  previous growth cycle. The pruning leaves a certain number of nodes that are able to guide the ideal direction of future growth of the vine towards a desired vine balance \cite{coombe1992viticulture, poni2016mechanical}. The aim of this work is that pruning which currently  requires experienced and skilled technicians, can eventually be conducted by intelligent and autonomous robotic systems \cite{botterill2017robot,GuadagnaDNN}.

The  proposed hierarchical control framework was validated through grapevine winter pruning experiments. The subsections below describe the details of our system specification and experimental results with the {Rolling Panda}. 

\begin{figure}[t] %H为当前位置，!htb为忽略美学标准，htbp为浮动图形
\centering %图片居中
\includegraphics[width=0.5\textwidth]{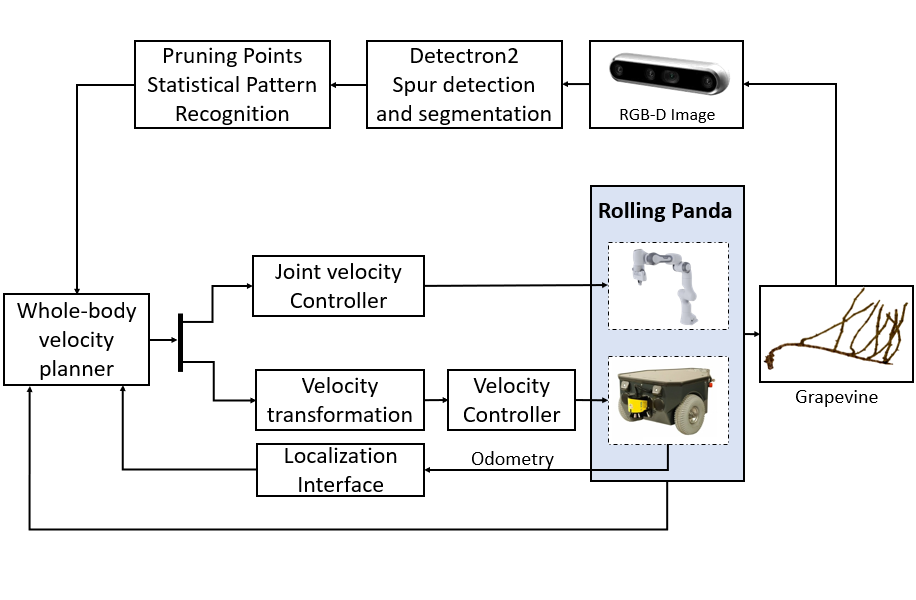} %插入图片，[]中设置图片大小，{}中是图片文件名
\caption{Overall paradigm of grapevine pruning experiment} %最终文档中希望显示的图片标题
\label{Fig.main2} %用于文内引用的标签
\end{figure}

\subsection{ Grapevine Pruning Task}

\begin{figure}[h] %H为当前位置，!htb为忽略美学标准，htbp为浮动图形
\centering %图片居中
\includegraphics[width=0.5\textwidth]{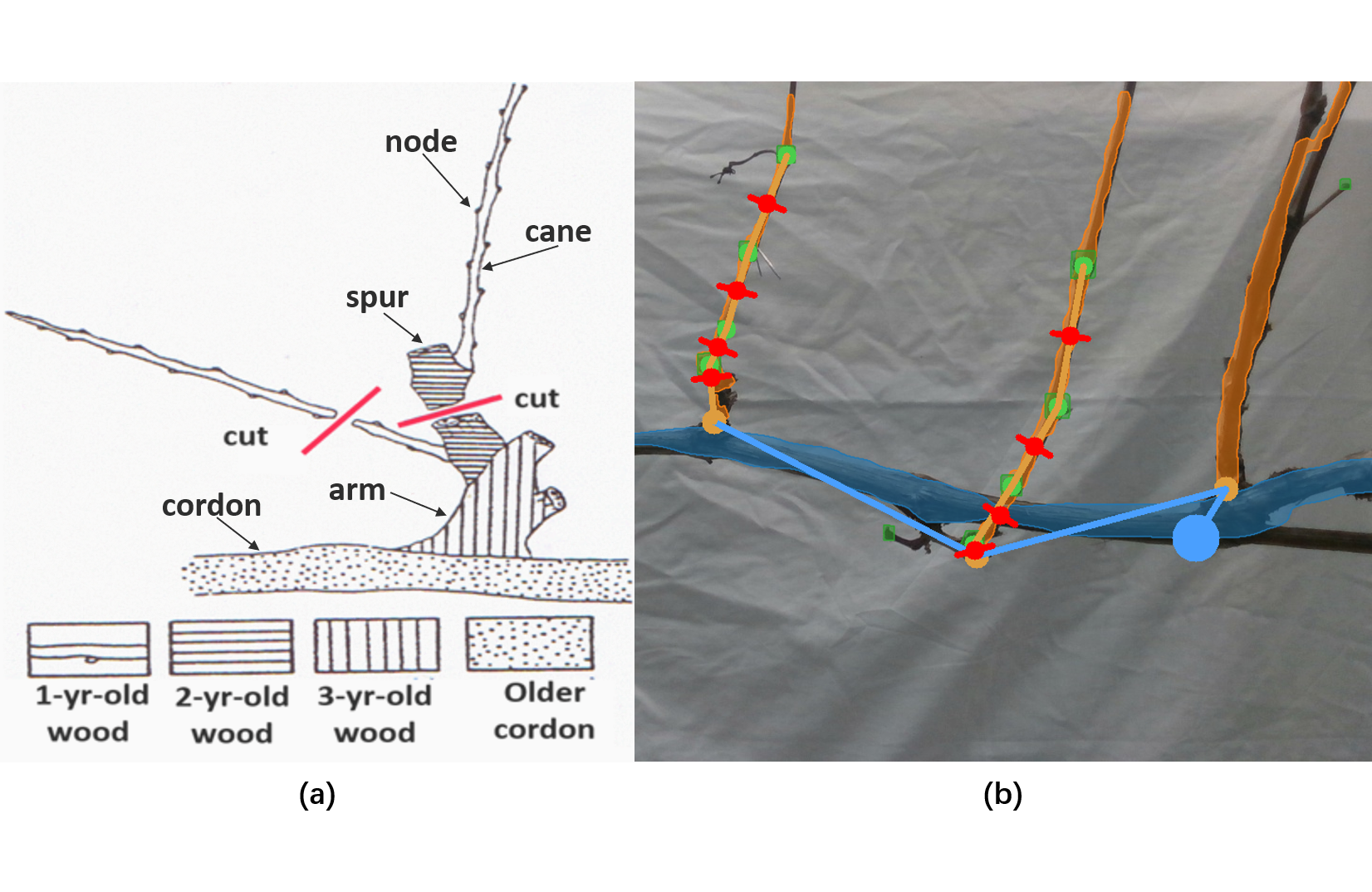} %插入图片，[]中设置图片大小，{}中是图片文件名
\caption{Effective pruning point illustration: (a) diagrammatic representation of pruning points from an agricultural and physiological perspective, (b) potential pruning points generated by deep learning } %最终文档中希望显示的图片标题
\label{pruning point illustration} %用于文内引用的标签
\end{figure}

The definition and determination of the  “correct” grapevine pruning point is based on  grapevine  modeling and physiological response \cite{perez2017image}.  Fig. \ref{pruning point illustration} (a) gives  a intuitive  and graphical illustration of the potential optimal pruning point.

In this paper, we use the visual perception methodology mentioned in \cite{hosseini2017designing} to determine potential pruning points, first creating a representative model of a grapevine plant using object segmentation on grapevine images and , second, generating a set of potential pruning  points.

%\begin{figure}[!htb] %H为当前位置，!htb为忽略美学标准，htbp为浮动图形
%\centering %图片居中
%\includegraphics[width=0.5\textwidth]{figures/output00425_inf_pp2.png} %插入图片，[]中设置图片大小，{}中是图片文件名
%\caption{Pruning point generation} %最终文档中希望显示的图片标题
%\label{Fig.main2} %用于文内引用的标签
%\end{figure}

\begin{figure*}[htbp] %H为当前位置，!htb为忽略美学标准，htbp为浮动图形
\centering %图片居中
\includegraphics[width=1\textwidth]{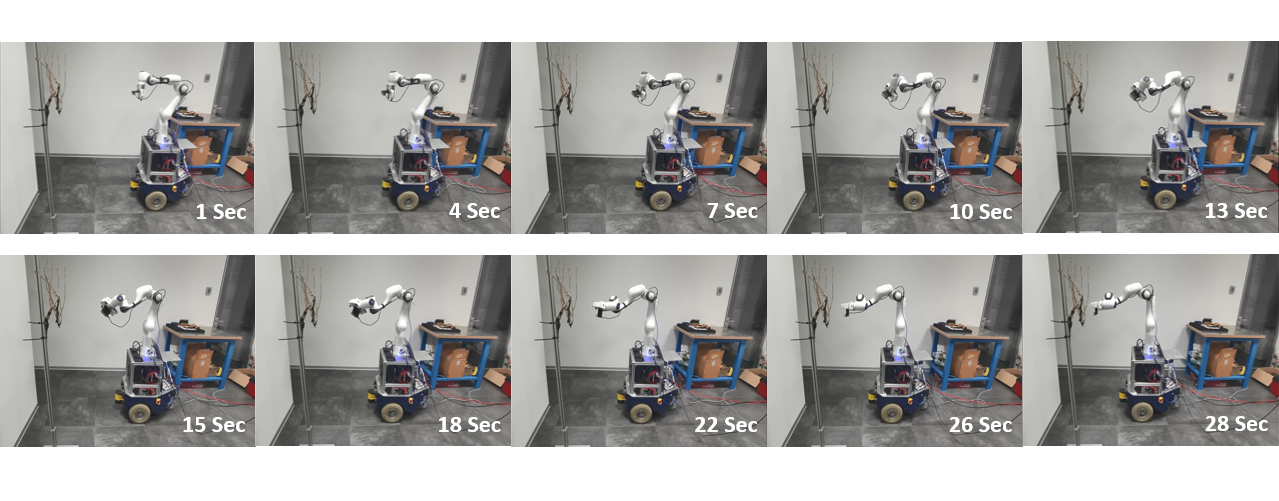} %插入图片，[]中设置图片大小，{}中是图片文件名
\caption{Experimental results for grapevine pruning point approach: snapshots} %最终文档中希望显示的图片标题
\label{results} %用于文内引用的标签
\end{figure*}

%In Fig. 9 (a), basic task and target basic task trajectory for manipulator are shown. Orientation of the basic task is not given here as in this experiment it is fixed when grasping. Like the constraint task, we use linear interpolation to get target basic task trajectory. From t=15.7 to t=18.1s, the manipulator is approaching the target object and from t=18.1s to t=18.5s, the gripper grips it. Fig. 9 (b) shows the basic task error for manipulator. Overshoot also occurs when the manipulator starts moving. Combined with Fig. 8 (b), we can see that in coordinated motion procedure, constraint task error reaches its peak because of the coupling between mobile base and mainipulator. However, the manipulator still tracks the desired trajectory well, especially in the period from t=18.1s to t=18.5s during which the gripper grips the object, the tracking error for basic task can be controlled less than 0.01m.

After the perception system finds the position of the pruning point in the global frame as showed in Fig. \ref{pruning point illustration} (b). The whole-body controller  generates a trajectory to approach the target pruning point using  quintic polynomial interpolation programming \cite{guan2005robotic}. The planned trajectory is  defined as follows: 

\begin{equation}\begin{array}{l}
P_x(t)=c_{0}+c_{1} t+c_{2} t^{2}+c_{3} t^{3}+c_{4} t^{4}+c_{5} t^{5} \\
P_v(t)=c_{1} t+2 c_{2} t+3 c_{3} t^{2}+4 c_{4} t^{3}+5 c_{5} t^{4} \\
P_a(t)=2 c_{2}+6 c_{3} t+12 c_{4} t^{2}+20 c_{5} t^{3}
\end{array}\end{equation}
where  $P_x$, $P_v$ and $P_a$ respectively correspond to the position, velocity, and acceleration in Cartesian space. Therefore, Eq. 13 can be rewritten as:

%$q_{m}=\left[\begin{array}{lll}x_{m} & y_{m} & \phi\end{array}\right]^{T} \in \mathbb{R}^{3}$

\begin{equation}\left[\begin{array}{cccccc}
1 & t_{s} & t_{s}^{2} & t_{s}^{3} & t_{s}^{4} & t_{s}^{5} \\
1 & t_{e} & t_{s}^{2} & t_{s}^{3} & t_{s}^{4} & t_{s}^{5} \\
0 & 1 & 2 t s & 3 t_{s}^{2} & 4 t_{s}^{3} & 5 t_{s}^{4} \\
0 & 1 & 2 t_{e} & 3 t_{e}^{2} & 4 t_{e}^{3} & 5 t_{e}^{4} \\
0 & 0 & 2 & 6 t_{s} & 12 t_{s}^{2} & 20 t_{s}^{3} \\
0 & 0 & 2 & 6 t_{e} & 12 t_{e}^{2} & 20 t_{e}^{3}
\end{array}\right]\left[\begin{array}{l}
c_{0} \\
c_{1} \\
c_{2} \\
c_{3} \\
c_{4} \\
c_{5}
\end{array}\right]=\left[\begin{array}{l}
x_{s} \\
x_{e} \\
v_{s} \\
v_{e} \\
a_{s} \\
a_{e}
\end{array}\right]\end{equation}
where $x_{s}$, $v_{s}$, and $a_{s}$ correspond to the position, velocity, and acceleration of the interpolation initial  point respectively, and  $x_{e}$, $v_{e}$ and $a_{e}$ correspond to the position, velocity, and acceleration of the interpolation end  point respectively.

\textit{Constraint task 1}: Singularities are configurations in which a robot loses control in one or more directions, and they should be avoided when planning and controlling robot motion. The manipulability calculation has been widely used and proven to be an effective way to keep robots away from singular configurations among several methods. Since singularities have no effect on the location of the mobile base, only upper manipulator configurations will be considered in this hierarchical control formulation. The term "manipulation" refers to the ability to manipulate something. The manipulability measurement is defined as

\begin{equation}
\omega\left(\xi_{\mathrm{c1}}\right)=\sqrt{\operatorname{det}\left({J}_{\xi}\left(\xi_{\mathrm{c1}}\right) {J}_{\xi}^{\mathrm{T}}\left(\xi_{\mathrm{c1}}\right)\right)}
\end{equation}

When the manipulability measurement is increased, the manipulator will move away from singularities. The following equation can be used to determine the corresponding joint inputs:
\begin{equation}
\dot{\xi}_{\mathrm{c1}}=k_{0}\left(\frac{\partial \omega\left(\xi_{\mathrm{c1}}\right)}{\partial \xi_{\mathrm{c1}}}\right)^{\mathrm{T}}
\end{equation}
where $k_0$ is a positive gain.

\textit{Constraint task 2}: Joint limits are physical constraints on robots that must be carefully considered in order to avoid damaging the robotic system. Certainly, only the upper manipulator joints have physical limitations. All joint angles are restricted to a range of $-180^{\circ}$ to $180^{\circ}$. To avoid joint limits, the artificial potential field technique \cite{khatib1985real} is used to computes the distance between the $i$th joint and its limits.

\begin{equation}
d_{i}=\min \left(\left\|\gamma_{i}-\gamma_{li}\right\|,\left\|\gamma_{i}-\gamma_{\mathrm{ui}}\right\|\right),
\end{equation}
where $\gamma_{i}$ denotes joint angle of the $i$-th joint, $\gamma_{11}$ and $\gamma_{\text {ul }}$ denote the lower and upper joint limits of the $i$-th joint, respectively. 

The $i$-th joint's "repulsive velocity" is defined as follows:
\begin{equation}
\dot{\xi}_{i, c2}=\left\{\begin{array}{ll}
k_{i} d_{i}^{2}, & d_{i} \leqslant \gamma_{\text {start }} \\
0, & d_{i}>\gamma_{\text {start }}
\end{array}\right.
\end{equation}
where $k_{i}$ denotes  a positive gain, and $\gamma_{\text {start }}$ denotes  the minimum distance to be free of repulsive force.

So far, we have modeled the main task (Rolling panda's end effector approach grapevine pruning point ) and low-level tasks (constraint task 1 and constraint task 2) at the velocity level. The application of mull-space projection technology is able to project constrained tasks to the null space of  main task, so that the robot can perform these tasks simultaneously and ensure the priority of the tasks.

\subsection{Results and Discussion}

%To better evaluate the reliability and robustness of the proposed hierarchical control formulation. 
The proposed controller for the non-holonomoic mobile manipulator can deal with grapevine pruning tasks while satisfying singularity avoidance and joint limitation avoidance, as shown in  Fig. \ref{results}.  The main task is to move the end-effector along the direction of $x$-axis to approach the pruning point and then rotate around $pitch$-axis to match the orientation of the pruning point. When the target pruning point is generated by the visual perception system, the whole-body controller can control the overall coordinated movement of the robot to reach the pruning point smoothly.

\section{Conclusions}
In this paper, which addresses the problem of grapevine winter pruning, we present a task-priority coordinated whole-body  motion controller for a non-holonomic mobile manipulator. The controller can plan and schedule  whole-body  coordinated motion  to complete the grapevine pruning task. The top priority task can be executed by employing all capabilities  (manipulation and locomotion)  of the robotic system. The second (lower) priority task is then projected into the null space of the top priority task, hence they have no impact on its execution. We conducted grapevine pruning experiment to demonstrate the performance of the proposed framework  using a two-wheeled mobile base with a 7-DoF robot manipulator. Our future work will involve the extension of the proposed framework for whole-body motion and perception coupling.

\section*{ACKNOWLEDGMENT}

We thank Sunny Katyara (Istituto Italiano di Tecnologia), Carlo Rizzardo (Istituto Italiano di Tecnologia) and Antonello Scaldaferri (Istituto Italiano di Tecnologia) for their assistance in experiment setup.

\bibliographystyle{IEEEtran}

\bibliography{reference}

\end{document}